\documentclass[11pt]{article}

\usepackage[a4paper,total={6.5in,9.5in}]{geometry}
\usepackage[T1]{fontenc}
\usepackage{libertine}
\usepackage{optidef}
\usepackage{amsmath}
\usepackage{amsthm}
\usepackage{amssymb}
\usepackage{mathtools}
\usepackage{algorithm}
\usepackage{algpseudocode}
\usepackage{float}
\usepackage{bbm}
\usepackage{listings}
\usepackage{booktabs}
\usepackage{comment}
\usepackage{subcaption}
\usepackage{enumitem}

\usepackage{hyperref} 
\hypersetup{
    linkbordercolor={1 0 0},
    pdftitle={UXps},
    pdfpagemode=FullScreen,
}

\usepackage[capitalize]{cleveref}
\usepackage[
backend=biber,
style=alphabetic,
]{biblatex}

\newcommand{\citep}{\parencite}
\newcommand{\citet}{\textcite}
\newtheorem{theorem}{Theorem}[section]
\newtheorem{lemma}{Lemma}[section]
\newtheorem{corollary}{Corollary}[section]
\newtheorem{definition}{Definition}[section]

\title{Sufficient and Necessary Explanations \\(and What Lies in Between)}

\author{Beepul Bharti\\\texttt{bbharti1@jhu.edu}}
\author{
    Beepul~Bharti$^{1,2}$\\
    {\small bbharti1@jhu.edu}
    \and
    Paul H. Yi~$^{3}$\\
    {\small paul.yi@stjude.org}
    \and
    Jeremias~Sulam$^{1,2,4}$\\
    \small {jsulam1@jhu.edu}
}
\date{
    \centering
    \small
    \vspace{1em}
    $^1$Mathematical Institute for Data Science (MINDS), Johns Hopkins University\\
    $^2$Department of Biomedical Engineering, Johns Hopkins University\\
    $^3$St. Jude Children's Research Hospital\\
    $^4$Department of Computer Science, Johns Hopkins University\\
    \vspace{1em}
    \today
}
\def\calD{{\mathcal D}}
\def\calX{{\mathcal X}}
\def\calY{{\mathcal Y}}
\def\calV{{\mathcal V}}
\def\calN{{\mathcal N}}

\def\R{{\mathbb R}}
\def\E{{\mathbb E}}

\def\X{{\bf X}}

\def\x{{\bf x}}

\def\cS{{S_c}}

\def\shap{{\text{shap}}}
\def\suf{{\text{suf}}}
\def\nec{{\text{nec}}}
\def\uni{{\text{uni}}}

\begin{document}
\maketitle
\begin{abstract}
    As complex machine learning models continue to find applications in high-stakes decision making scenarios, it is crucial that we can explain and understand their predictions. Post-hoc explanation methods provide useful insights by identifying important features in an input $x$ with respect to the model output $f(\x)$. In this work we formalize and study two precise notions of feature importance for general machine learning models: \emph{sufficiency} and \emph{necessity}. We demonstrate how these two types of explanations, albeit intuitive and simple, can fall short in providing a complete picture of which features a model finds important. To this end, we propose a unified notion of importance that circumvents these limitations by exploring a continuum along a necessity-sufficiency axis. Our unified notion, we show, has strong ties to other popular definitions of feature importance, like those based on conditional independence and game-theoretic quantities like Shapley values. Crucially, we demonstrate how a unified perspective allows us to detect important features that could be missed by either of the previous approaches alone.
\end{abstract}

\newpage
\tableofcontents
\newpage

\section{Introduction}
\label{sec:intro}
Over recent years, modern machine learning (ML) models, mostly deep learning based, have achieved impressive results across several complex domains. We now have models that can solve difficult image classification, inpainting, and segmentation problems, perform accurate text and sentiment analysis, predict the three-dimensional conformation of proteins, and more \citep{lecun2015deep, wang2023nature}. Despite their success, the rapid integration of these models into society requires caution due to their complexity and unintelligibility \citep{executiveorder2023}. Modern ML systems are, by and large black-boxes, consisting of millions of parameters and non-linearities that obscure their prediction-making mechanisms from users, developers, and auditors. This lack of clarity raises concerns about explainability, transparency, and accountability \citep{zednik2021solving, tomsett2018interpretable}. Thus, understanding how these models work is essential for their safe deployment.

The lack of explainability has spurred research efforts in eXplainable AI (XAI), with a major focus on developing post-hoc methods to explain black-box model predictions, especially at a \emph{local} level. For a model $f$ and input $\x \in \R^d$, these methods aim to identify which features in $\x$ are \emph{important} for the model's prediction, $f(\x)$. They do so by estimating a notion of importance for each feature (or groups) which allows for a ranking of importance. Popular methods include CAM \citep{zhou2016learning}, LIME \citep{ribeiro2016should}, gradient-based approaches \citep{selvaraju2017grad, shrikumar2017learning, jiang2021layercam}, rate-distortion techniques \citep{kolek2021cartoon, kolek2022rate}, Shapley value-based explanations \citep{chen2018shapley, teneggi2022fast, mosca2022shapbased}, perturbation-based methods \citep{fong2017interpretable, fong2019understanding, dabkowski2017real}, among others \citep{chen2018learning, yoon2018invase, jethani2021have, wang2021probabilistic, ribeiro2018anchors}. However, many of these approaches lack rigor, as the meaning of their computed scores is often ambiguous. For example, it's not always clear what large or negative gradients signify or what high Shapley values reveal about feature importance. To address these concerns, other research has focused on developing explanation methods based on logic-based definitions \citep{ignatiev2020contrastive, darwiche2020reasons, darwiche2022computation, shih2018symbolic}, conditional hypothesis testing \cite{teneggi2023shap,tansey2022holdout}, among formal notions. While these methods are a step towards rigor, they have drawbacks, including reliance on complex automated reasoners and limited ability to communicate their results in an understandable way for human decision-makers.

In this work, we advance XAI research by formalizing rigorous mathematical definitions and approaches, grounded in the intuitive concepts of \emph{sufficiency} and \emph{necessity}, to explain complex ML models. We begin by illustrating how sufficient and necessary explanations offer valuable, albeit incomplete, insights into feature importance. To address this issue, we propose and study a more general unified framework for explaining models. We offer two novel perspectives on our framework through the lens of conditional independence and Shapley values, and crucially, show how it reveals new insights into feature importance.

\subsection{Summary of our Contributions}
We study two key notions of importance: sufficiency and necessity, both which evaluate the importance of a set of features in $\x$, with respect to the prediction $f(\x)$, of an ML model. A sufficient set of features preserves the model's output, while a necessary set, when removed or perturbed, renders the output uninformative. Although sufficiency and necessity appear complementary, their precise relationship remains unclear. When do sufficient and necessary subsets overlap or differ? When should we prioritize one over the other, or seek features that are both necessary \textit{and} sufficient? To address these questions, we analyze sufficiency and necessity and propose a \emph{unification} of both. Our contributions are summarized as follows:

\begin{enumerate}
    \item We formalize precise mathematical definitions of sufficient and necessary model explanations for arbitrary ML predictors.
    \item We propose a unified approach that combines sufficiency and necessity, analyzing their relationships and exploring exploring when and how they align or differ. Furthermore, we reveal its strong ties to conditional independence and Shapley values, a game-theoretic measure of feature importance.
    \item Through experiments of increasing complexity, illustrate how our unified perspective can reveal new, important, and more complete insights into feature importance.
    \vspace{-0.2cm}
\end{enumerate}

\section{Sufficiency and Necessity}
\label{sec:suff_vs_necc}
{\bf Notation.} We use boldface uppercase letters to denote random vectors (e.g., $\X$) and lowercase letters for their values (e.g., $\x$). For a subset $S \subseteq [d] \coloneqq \{1, \dots, d\}$, we denote its cardinality by $|S|$ and its complement $\cS = [d] \setminus S$. Subscripts index features; e.g., the vector $\x_S$ represents the restriction of $\x$ to the entries indexed by $S$. 

{\bf Setting.} We consider a supervised learning setting with an unknown distribution $\calD$ over $\calX \times \calY$, where $\calX \subseteq \mathbb{R}^d$ is a $d$-dimensional feature space and $\calY \subseteq \mathbb{R}$ is the label space. We assume access to a predictor $f: \calX \mapsto \calY$ that was trained on samples from $\calD$. For an input \(\mathbf{x} = (x_1, \dots, x_d) \in \mathbb{R}^d\), our goal is to identify the important features of \(\mathbf{x}\) for the prediction \(f(\mathbf{x})\). To define feature importance precisely, we use the \emph{average restricted prediction},
\begin{align}
f_S(\mathbf{x}) = \underset{\mathbf{X}_{\mathcal{S}} \sim \mathcal{V}_{\mathcal{S}}}{\mathbb{E}}[f(\mathbf{x}_S, \mathbf{X}_{\mathcal{S}})]
\end{align}
where \(\mathbf{x}_S\) is fixed, and \(\mathbf{X}_{\mathcal{S}}\) is a random vector drawn from an arbitrary reference distribution \(\mathcal{V}_{\mathcal{S}}\). This strategy, popularized in \citep{lundberg2017unified, lundberg2020local}, allows us to query a predictor \(f\) that accepts only \(d\)-dimensional inputs and analyze its behavior when specific sets of features in \(\mathbf{x}\) are retained or removed.

\subsection{Definitions}
We introduce two intuitive notions to quantify the importance of a subset $S$ for a prediction $f(\x)$. For a model $f$, we begin by evaluating its baseline behavior over an arbitrary reference distribution $\calV$: $f_{\emptyset}(\x) = {\E}[f(\X_{[d]})]$. Then, for a sample $\x$ and its prediction $f(\x)$, we can pose two simple questions: 
\begin{center}
    \emph{Which set of features, $S$, satisfies $f_S(\x) \approx f(\x)$ or $f_{\cS}(\x) \approx f_{\emptyset}(\x)$}?
\end{center} 
These questions explores the sufficiency and necessity of subset $S$, which we define formally in the following definitions

\begin{definition}[Sufficiency] Let $\epsilon \geq 0$ and let $\rho: \R \times \R \mapsto \R$ be a metric on $\R$. A subset $S \subseteq [d]$ is $\epsilon$-sufficient with respect to a distribution $\calV$ for $f$ at $\x$ if
\begin{equation}
\Delta^{\suf}_\calV(S, f, \x) \triangleq \rho(f(\x), f_S(\x)) \leq \epsilon.
\end{equation}
Furthermore, $S$ is $\epsilon$-super sufficient if all supersets $\widetilde{S} \supseteq S$ are $\epsilon$-sufficient.
\end{definition}

This notion of sufficiency is straightforward: a subset $S$ is $\epsilon$-sufficient with respect to a reference distribution $\calV$ if, with $\x_S$ fixed, the average restricted prediction $f_S(\x)$ is within $\epsilon$ from the original prediction $f(\x)$. This is further strengthened by super-sufficiency: a subset $S$ is $\epsilon$-super sufficient if $ \rho(f(\x), f_S(\x)) \leq \epsilon$ and, for any superset $\widetilde{S}$ of $S$, $\rho(f(\x), f_{\widetilde{S}}(\x)) \leq \epsilon$. This simply means including more features in $S$ still keeps $f_S(\x)$ $\epsilon$ close to $f(\x)$. To find a small sufficient subset $S$ of small cardinality $\tau>0$, we can solve the following optimization problem:
\begin{argmini*}|l|[2]
    {S \subseteq [d]}
    {\Delta^{\suf}_\calV(S, f, \x) \quad \text{subject to} \quad |S| \leq \tau.}
    {}
    {\tag{P$_{\suf}$} {\label{eq:suff_opt}}}
\end{argmini*}
We will refer to this problem as the \emph{sufficiency problem}, or \eqref{eq:suff_opt}. Using analogous ideas, we also define necessity and formulate an optimization problem to find small necessary subsets.

\begin{definition}[Necessity] Let $\epsilon \geq 0$ and denote $\rho: \R \times \R \mapsto \R$ to be metric on $\R$. A subset $S \subseteq [d]$ is $\epsilon$-necessary with respect to a distribution $\calV$ for $f$ at $\x$ if
\begin{equation}
    \Delta^{\nec}_\calV(S, f, \x)  \triangleq \rho(f_{\cS}(\x), f_{\emptyset}(\x)) \leq \epsilon.
\end{equation}
Furthermore, $S$ is $\epsilon$-super necessary if all supersets $\widetilde{S} \supseteq S$ are $\epsilon$-necessary.
\end{definition}

Here, a subset $S$ is $\epsilon$-necessary if marginalizing out the features in $S$ with respect to the distribution $\calV_{S}$, results in an average restricted prediction $f_{\cS}(\x)$ that is $\epsilon$ close to $f_{\emptyset}(\x)$ -- the average baseline prediction of $f$ over $\calV_{[d]}$. Furthermore, $S$ is $\epsilon$-super necessary if $\rho(f_S(\x), f(\x)) \leq \epsilon$ and any superset $\widetilde{S}$ of $S$ is $\epsilon$-necessary. To identify a $\epsilon$-necessary subset $S$ of small cardinality $\tau > 0$, one can solve the following optimization problem, which we refer to as the \emph{necessity} problem or \eqref{eq:necc_opt}.  
\begin{argmini*}|l|[2]
    {S \subseteq [d]}
    {\Delta^{\nec}_\calV(S, f, \x)  \quad \text{subject to} \quad |S| \leq \tau} 
    {}
    {\tag{P$_{\nec}$} {\label{eq:necc_opt}}}.
\end{argmini*}

\section{Related Work}
\label{sec:related_work}
Notions of sufficiency, necessity, the duality between the two, and their connections with other feature attribution methods have  been studied to varying degrees in XAI research. We comment on the main related works in this section.

{\bf Sufficiency.}  The notion of sufficient features has gained significant attention in recent research. \citet{shih2018symbolic} explore a symbolic approach to explain Bayesian network classifiers and introduce prime implicant explanations, which are minimal subsets $S$ that make features in the complement irrelevant to the prediction $f(\x)$. For models represented by a finite set of first-order logic (FOL) sentences, \citet{ignatiev2020contrastive} refer to prime implicants as abductive explanations (AXp's). For classifiers defined by propositional formulas and inputs with discrete features, \citet{darwiche2020reasons} refer to prime implicants as sufficient reasons and define a complete reason to be the disjunction of all sufficient reasons. They present efficient algorithms, leveraging Boolean circuits, to compute sufficient and complete reasons and demonstrate their use in identifying classifier dependence on protected features that should not inform decisions. For more complex models, \citet{ribeiro2018anchors} propose high-precision probabilistic explanations called anchors, which represent local, sufficient conditions. For $\x$ positively classified by $f$, \citet{wang2021probabilistic} propose a greedy approach to solve \eqref{eq:suff_opt} while the preservation method by \citet{fong2017interpretable} relaxes $S$ to $[0,1]^d$.

{\bf Necessity.} There has also been significant focus on identifying necessary features -- those that, when altered, lead to a change in the prediction $f(\x)$. For models expressible by FOL sentences, \citet{ignatiev2019relating} define prime implicates as the minimal subsets that when changed, modify the prediction $f(\x)$ and relate these to adversarial examples. For Boolean models predicting on samples $\x$ with discrete features, \citet{ignatiev2020contrastive} and \citep{darwiche2020reasons}  refer to prime implicates as contrastive explanations (CXp's) and necessary reasons, respectively. Beyond boolean functions, for $\x$ positively classified by a classifier $f$, \citet{fong2019understanding} relax $S$ to $[0,1]^d$ and propose the deletion method to approximately solve \eqref{eq:necc_opt}.

{\bf Duality Between Sufficiency and Necessity.} \citet{dabkowski2017real} characterize the preservation and deletion methods as discovering the \emph{smallest sufficient} and \emph{destroying region} (SSR and SDR). They propose combining the two but do not explore how solutions to this approach may differ from individual SSR and SDR solutions. \citet{ignatiev2020contrastive} show that AXp's and CXp's are minimal hitting sets of another by using a hitting set duality result between minimal unsatisfiable and correction subsets. The result enables the identification of AXp's from CXp's and vice versa.

{\bf Sufficiency, Necessity, and General Feature Attribution Methods.} Precise connections between sufficiency, necessity, and other popular feature attribution methods (such as Shapley values \citep{shapley1951notes, chen2018shapley, lundberg2017unified}) remains unclear. To our knowledge, \citet{covert2021explaining} provide the only work examining these approaches \citep{fong2017interpretable, fong2019understanding, dabkowski2017real} in the context of general removal-based methods, i.e., methods that remove certain input features to evaluate different notions of importance. The work of \citet{watson21a} is also relevant to our work,  as it formalizes a connection between notions of sufficiency and Shapley values. With the specific payoff function \footnote{Payoff functions are an instrumental tool in game-theoretic approaches. See further \cref{sec:Shapley} for further details.} defined as $v(S) = \mathbb{E}[f(\mathbf{x}_{S}, \mathbf{X}_{\cS})]$, they show how each summand in the Shapley value measures the sufficiency of feature $i$ to a particular subset.

\section{Unifying Sufficiency and Necessity}
\label{sec:unifying_suff_and_necc}
Given a model $f$, sample $\x$, and reference distribution $\calV$, we can identify a small set of important features $S$ by solving either \eqref{eq:suff_opt} or \eqref{eq:necc_opt} \footnote{Solving \eqref{eq:suff_opt} or \eqref{eq:necc_opt} is NP-hard for general non-convex functions $f$. We do not concern ourselves with the computational efficiency of these problems as there exist tractable relaxations \citep{kolek2021cartoon, kolek2022rate}.}. 
While both methods are popular \citep{kolek2021cartoon, kolek2022rate, fong2017interpretable, bhalla2023verifiable, yoon2018invase}, simply identifying a small sufficient or necessary subset may not provide a complete picture of how $f$ uses $\x$ to make a prediction. To see why, consider the following scenario: for a fixed $\tau > 0$, let $S^*$ be a $\epsilon$-sufficient solution to \eqref{eq:suff_opt}, so that $|S^*| \leq \tau$ and 
\begin{align}
    \Delta^{\suf}_\calV(S, f, \x) \leq \epsilon.
\end{align}
While $S^*$ is $\epsilon$-sufficient, it can also be true that 
\begin{align}
    \Delta^{\nec}_\calV(S, f, \x)>\epsilon
\end{align} 
indicating $S^*$ is {\bf not} $\epsilon$-necessary: indeed, this can simply happen when its complement, $S_c^*$, contains important features. This scenario raises two questions:
\begin{enumerate}
    \item How different are sufficient and necessary features?
    \item How does varying the levels of sufficiency and necessity affect the optimal set of important features?
\end{enumerate}

In order to provide answers to these questions (and to avoid the scenario above) we propose to search for a small set $S$ that is both sufficient \emph{and} necessary by combining problems \eqref{eq:suff_opt} and \eqref{eq:necc_opt}. Consider $\Delta^{\uni}_\calV(S, f, \x, \alpha)$, a convex combination of both $\Delta^{\suf}_\calV(S, f, \x)$ and $\Delta^{\nec}_\calV(S, f, \x)$
\begin{align}
    \Delta^{\uni}_\calV(S, f, \x, \alpha) = \alpha\cdot \Delta^{\suf}_\calV(S, f, \x) + (1-\alpha)\cdot\Delta^{\nec}_\calV(S, f, \x)
\end{align}
where $\alpha \in [0,1]$ controls the extent to which $S$ is required to be sufficient vs. necessary. Our \emph{unified problem}, \eqref{eq:uni_opt}, can be expressed as:
\begin{argmini*}|l|[2]
    {S \subseteq [d]}
    {\Delta^{\uni}_\calV(S, f, \x, \alpha) \quad \text{subject to} \quad |S| \leq \tau}
    {}
    {\tag{P$_{\uni}$} {\label{eq:uni_opt}}}.
\end{argmini*}
When $\alpha$ is $1$ or $0$, $\Delta^{\uni}_\calV(S, f, \x, \alpha)$ reduces to $\Delta^{\suf}_\calV(S, f, \x)$ or $\Delta^{\nec}_\calV(S, f, \x)$, respectively. In these extreme cases, $S$ is only sufficient or necessary.  In the remainder of this work we will theoretically analyze \eqref{eq:uni_opt}, characterize its solutions, and provide different interpretations of what properties the solutions have through the lens of conditional independence and game theory. In the experimental section, we will show that solutions to \eqref{eq:uni_opt} provide insights that neither \eqref{eq:suff_opt} nor \eqref{eq:necc_opt} offer.

\subsection{Solutions to the Unified Problem}
We begin with a simple lemma that demonstrates why \eqref{eq:uni_opt} enforces both sufficiency and necessity.
\begin{lemma} 
    \label{lemma:lemma1}
    Let $\alpha \in (0,1)$. For $\tau >0$, denote $S^*$ to be a solution to \eqref{eq:uni_opt} for which $\Delta^{\uni}_\calV(S, f, \x, \alpha)= \epsilon$. Then, $S^*$ is $\frac{\epsilon}{\alpha}$-sufficient and $\frac{\epsilon}{1-\alpha}$-necessary. Formally,
    \begin{align}
        0 \leq \Delta^{\suf}_\calV(S^*, f, \x) \leq \frac{\epsilon}{\alpha} \quad \text{and} \quad 0 \leq \Delta^{\nec}_\calV(S^*, f, \x) \leq \frac{\epsilon}{1-\alpha}.
    \end{align}
\end{lemma}
The proof of this result, and all others, is included \cref{supp:proofs}. This result illustrates that solutions to \eqref{eq:uni_opt} satisfy varying definitions of sufficiency and necessity. Furthermore, as $\alpha$ increases from 0 to 1, the solution shifts from being highly necessary to highly sufficient. In the following results, we will show \textit{when} and \textit{how} solutions to \eqref{eq:uni_opt} are similar (and different) to those of \eqref{eq:suff_opt} and \eqref{eq:necc_opt}. To start, we present the following lemma, which will be useful in subsequent results.
\begin{lemma} 
    \label{lemma:lemma2}
    For $0 \leq \epsilon < \frac{\rho(f(\x), f_{\emptyset}(\x))}{2}$, denote $S^*_{\suf}$ and $S^*_{\nec}$ to be $\epsilon$-sufficient and $\epsilon$-necessary sets. Then, if $S^*_{\suf}$ is $\epsilon$-super sufficient or $S^*_{\nec}$ is $\epsilon$-super necessary, we have
    \begin{align}
        S^*_{\suf} \cap S^*_{\nec} \neq \emptyset.
    \end{align}
\end{lemma}
This lemma demonstrates that, given \(\epsilon\)-sufficient and necessary sets \(S^*_{\suf}\) and \(S^*_{\nec}\), if either additionally satisfies the stronger notions of super sufficiency or necessity, they must share some features. This proves useful in characterizing a solution to \eqref{eq:uni_opt}, which we now do in the following theorem.
\begin{theorem}
\label{theorem:thm1}
Let $\tau_1, \tau_2 > 0$ and $0 \leq \epsilon < \frac{1}{2}\cdot\rho(f(\x), f_{\emptyset}(\x))$. Denote $S^*_{\suf}$ and $S^*_{\nec}$ to be $\epsilon$-super sufficient and $\epsilon$-super necessary solutions to \eqref{eq:suff_opt} and \eqref{eq:necc_opt}, respectively, such that $|S^*_{\suf}~| = \tau_1$ and $|S^*_{\nec}| = \tau_2$.
Then, there exists a set $S^*$ such that
\begin{align}
    \Delta^{\uni}_\calV(S^*, f, \x, \alpha) \leq \epsilon \quad \text{and} \quad
    \max(\tau_1, \tau_2) \leq |S^*| < \tau_1 + \tau_2.
\end{align}
Furthermore, if $S^*_{\suf} \subseteq S^*_{\nec}$ or $S^*_{\nec} \subseteq S^*_{\suf}$. then $S^* = S^*_{\nec}$ or $S^* = S^*_{\suf}$, respectively.
\end{theorem}
This result demonstrates that solutions to \eqref{eq:uni_opt}, \eqref{eq:suff_opt}, and \eqref{eq:necc_opt} can be closely related. As an example, consider features that are $\epsilon$-super sufficient, $S^*_{\suf}$. If we have domain knowledge that $S^*_{\suf}\subseteq S^*_{\nec}$, and $S^*_{\nec}$ is $\epsilon$-super necessary, then $S^*_{\nec}$ is in fact the solution to the \eqref{eq:uni_opt} problem. Conversely, if we know that $S^*_{\suf}$ is $\epsilon$-super necessary along with being a subset of $\epsilon$-super sufficient set $S^*_{\suf}$, then $S^*_{\suf}$ will be a solution to the \eqref{eq:uni_opt} problem.

\section{Two Perspectives of the Unified Approach}
\label{sec:perspectives}
In the previous section, we characterized solutions to \eqref{eq:uni_opt} and their connections to those of \eqref{eq:suff_opt} and \eqref{eq:necc_opt}. To better understand sufficiency, necessity, and their unification, we will provide two alternative perspectives of our unified framework through the lens of conditional independence and Shapley values.

\subsection{A Conditional Independence Perspective}
\label{sec:ci}
Here we demonstrate how our sufficiency, necessity, and our unified approach, can be understood as measuring conditional independence relations between features $\X$ and labels $Y$.
\begin{corollary}
    \label{corollary:corollary1}
    Suppose for any $S \subseteq [d]$, $\calV_S = p({\X}_{S} \mid {\X}_{\cS} = {\x}_{\cS})$. Let $\alpha \in (0,1)$, $\epsilon \geq 0$, and denote $\rho: \R \times \R \mapsto \R$ to be a metric on $\R$. Furthermore, for $f(\X) = \E[Y \mid \X]$ and $\tau >0$, let $S^*$ be a solution to \eqref{eq:uni_opt} such that $\Delta^{\uni}_\calV(S, f, \x, \alpha) = \epsilon$. Then, $S^*$ satisfies the following conditional independence relations,
        \begin{align}
            &\rho\left(\E[Y \mid \x],~ \E[Y \mid \X_{S^*} = \x_{S^*}]\right) \leq \frac{\epsilon}{\alpha} \quad \text{and} \quad \rho\left(\E[Y \mid \X_{S^*_c} = \x_{S^*_c}],~ \E[Y]\right) \leq \frac{\epsilon}{1-\alpha}.
        \end{align}
\end{corollary}
The assumption in this corollary is that, $\forall~S \subseteq [d]$, $f_S(\x)$ is evaluated using the conditional distribution $p(\X_{\cS} \mid {\X}_S = \x_S)$ as the reference distribution $\calV_S$. Given the recent advancements in generative models \citep{song2019generative, ho2020denoising, song2021scorebased}, this assumption is (approximately) reasonable in many practical settings, as we will demonstrate in our experiments. With this reference distribution, the results shows that for the model $f(\X) = \E[Y \mid \X]$ and a sample $\x$, the minimizer $S^*$ of \eqref{eq:uni_opt} approximately satisfies two conditional independence properties. First, $S^*$ is sufficient in that, when the features in $S^*$ are fixed, the complement, $\cS^*$, offers little-to-no additional information about $Y$. Second, $S^*$ is necessary because when we marginalize it out and rely only on the features in $\cS^*$, the information gained about $Y$ is minimal and similar to $\E[Y=1]$.

\subsection{A Shapley Value Perspective}
\label{sec:Shapley}
In the previous section, we detailed the conditional independence relations one gains from solving \eqref{eq:uni_opt}. We now present an arguably less intuitive result that shows that solving \eqref{eq:uni_opt} is equivalent to maximizing the lower bound of Shapley value. Before presenting our result, we provide a brief background on this game-theoretic quantity.

{\bf Shapley Values.} Shapley values use game theory to measure the importance of players in a game. Let the tuple $([n], v)$ represent a cooperative game with players $[n] = \{1, 2, \dots, n\}$ and denote a characteristic function $v(S): \mathcal{P}([n])\to\R$, which maps the power set of $[n]$ to the reals. Then, the Shapley value \citep{shapley1951notes} for player $j$ in the cooperative game $([n], v)$ is 
\begin{align}
    \phi^{\shap}_j([n], v) = \sum_{S \subseteq [n] \setminus \{j\}} w_S \cdot \left[v(S \cup \{j\}) - v(S)\right]
\end{align} 
where $w_S = \frac{|S|!(n - |S| - 1)!}{n!}$. The Shapley value is the only solution concept that satisfies the desirable axioms of additivity, nullity, symmetry, and linearity \citep{owen2013game}. In the context of XAI and feature importance, Shapley values are widely used to measure local feature importance by treating input features as players in a game \citep{covert2020understanding, teneggi2022fast, chen2018shapley, lundberg2017unified}. Given a sample $\x \in \R^d$ and a model $f$, the goal is to evaluate the importance of each feature $j \in [d]$ for the prediction $f(\x)$. This is done by defining a cooperative game $([d], v)$, where $v(S)$ is a characteristic function that quantifies how the features in $S$ contribute to the prediction. Different choices of $v(S)$ can be found in \citep{lundberg2017unified, sundararajan2020many, watson2024explaining}. Although computing $\phi^{\shap}_j([d], v)$ is computationally intractable, several practical methods for estimation have been developed \citep{chen2023algorithms, teneggi2022fast, zhang2023towards, lundberg2020local}. While Shapley values are popular across various domains \citep{moncada2021explainable, zoabi2021machine, liu2021evaluating}, few works, aside from \citet{watson21a}, explore their connections to sufficiency and necessity.

With this background, we now present our result. Recall solving \eqref{eq:uni_opt} obtains a small subset $S$ with low $\Delta^{\uni}_\calV(S, f, \x, \alpha)$. Notice that in \eqref{eq:uni_opt} there is a natural \emph{partitioning} of the features into two sets, $S$ and $\cS$. In the follow theorem we demonstrate that searching for a small subset $S$ with minimal $\Delta^{\uni}_\calV(S, f, \x, \alpha)$ is equivalent to maximizing a lower bound on the Shapley value in a two player game.

\begin{theorem}
\label{theorem:thm2}
Consider an input $\x$ for which $f(\x) \neq f_{\emptyset}(\x)$.  Denote by $\Lambda_d = \{S, \cS \}$ the partition of $[d] = \{1, 2, \dots, d\}$, and define the characteristic function to be $v(S) = -\rho(f(\x), f_{S}(\x))$. Then,
\begin{align}
   \phi^{\shap}_S(\Lambda_d, v) \geq \rho(f(\x), f_{\emptyset}(\x)) - \Delta^{\uni}_\calV(S, f, \x, \alpha).
\end{align}
\end{theorem}
This result has important implications. When the feature space is partitioned into 2 disjoint sets, $S$ and $\cS$, where each is a player in a cooperative game, then in searching for an $S$ with small $\Delta_{\uni}(S, \calV, \frac{1}{2})$, as we do in \eqref{eq:uni_opt}, we are searching for a player, $S$, with a large lower bound on its Shapley value. Note this connection we show is different from the one presented by \citep{watson21a}. They show the Shapley value of feature $i$ is a measure of this feature $i$'s sufficiency subsets $S \subseteq [d]$. In conclusion, our result provides a new and different and complementary interpretation to the sufficiency, necessity, and our proposed unified method through the lens of game theory. 
\vspace{-0.2cm}

\section{Experiments}
\label{sec:experiments}
We demonstrate our theoretical findings in multiple settings of increasingly complexity: two tabular data tasks (on synthetic data and the US adult income dataset \citep{ding2021retiring}) and two high-dimensional image classification tasks using the RSNA 2019 Brain CT Hemorrhage Challenge \citep{flanders2020construction} and CelebA-HQ datasets \citep{CelebAMask-HQ}.

\subsection{Tabular Data}
In the following examples, we analyze solutions to \eqref{eq:uni_opt} for varying levels of sufficiency vs. necessity and multiple size constraints. We learn a predictor $f$ and, for 100 new samples, solve \(\eqref{eq:uni_opt}\) for \(\tau \in \{3, 6, 9\}\) and \(\alpha \in [0,1]\), with \(\rho(a,b) = |a-b|\) and \(\calV_S = p({\X}_S \mid {\X}_{\cS} = {\bf x}_{\cS})\). For a fixed $\tau$ and sample \(\x\),  we denote \( S^*_{\alpha_i} \) to be a solution to \(\eqref{eq:uni_opt}\) for \(\alpha_i\). 
To analyze the stability of \( S^*_{\alpha_i} \) as sufficiency and necessity vary, we report the normalized average Hamming distance \citep{hamming1950error} between \( S^*_{\alpha_i} \) and \( S^*_{0} \), along with 95\% confidence intervals, as a function of \(\alpha\).

\subsubsection{Linear Regression}
We begin with a regression example. Features $\X$ are distributed according to $\cal{N}({\boldsymbol \mu}, {\bf AA^T})$ with $\boldsymbol\mu = \left[ 2^i\right]^{d}_{i=1}$ and ${\bf A}_{i,j} \sim U(0,1)$. The response is \( Y = \boldsymbol{\beta}^T\X + \boldsymbol{\epsilon} \), with \(\boldsymbol\beta = 32 \cdot [2^{-i}]_{i=1}^{d}\) and \(\boldsymbol{\epsilon} \sim \mathcal{N}(\mathbf{0}, \mathbf{I}_{d \times d})\). With \(d = 10\) our model is \( f(\X) = \hat{\boldsymbol\beta}^T\X \), where \(\hat{\boldsymbol\beta}\) is the least squares solution.

{\bf Stability of Unified Solutions.} \cref{fig:regression} shows that when solutions are constrained to be small (\(\tau = 3\)), increasing \(\alpha\) to enforce greater sufficiency results in a steady increase in Hamming distance,  indicating that the solutions \( S^*_{\alpha_i} \) are consistently changing. When larger solutions are allowed (\(\tau = 6\)), \( S^*_{\alpha_i} \) rapidly changes with the introduction of sufficiency, as seen by the initial steep rise in Hamming distance. However, as $\alpha$ continues to increase, this distance grows more gradually. Lastly, when the solution size approaches the dimensionality of the feature space (\(\tau = 9\)), small to intermediate levels of sufficiency do not significantly alter the solutions. However, requiring high levels of sufficiency (\(\alpha > 0.8\)) leads to extreme changes in the solutions, as shown by a sharp increase in Hamming distance. 

\subsubsection{American Community Survey Income (ACSIncome)}
We use the ACSIncome dataset for California, including 10 demographic and socioeconomic features such as age, education, occupation, and geographic region. We train a Random Forest classifier to predict whether an individual's annual income exceeds \$50K, achieving a test accuracy $\approx 81\%$. 
\begin{figure*}
    \centering
    \begin{subfigure}{0.45\textwidth}
        \centering
        \includegraphics[width=\linewidth]{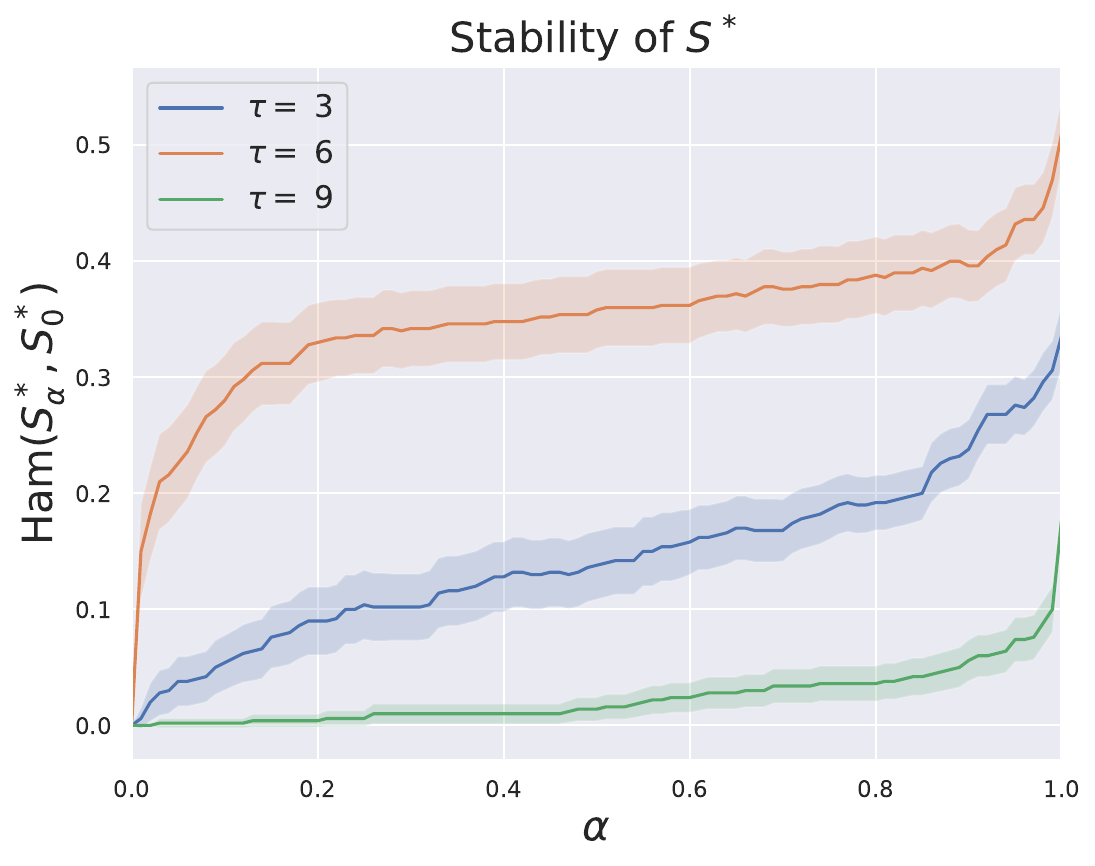}
        \caption{\hspace{0.2em}Regression}
        \label{fig:regression}
    \end{subfigure}
    \hspace{0.75cm}
    \begin{subfigure}{0.45\textwidth}
        \centering
        \includegraphics[width=\linewidth]{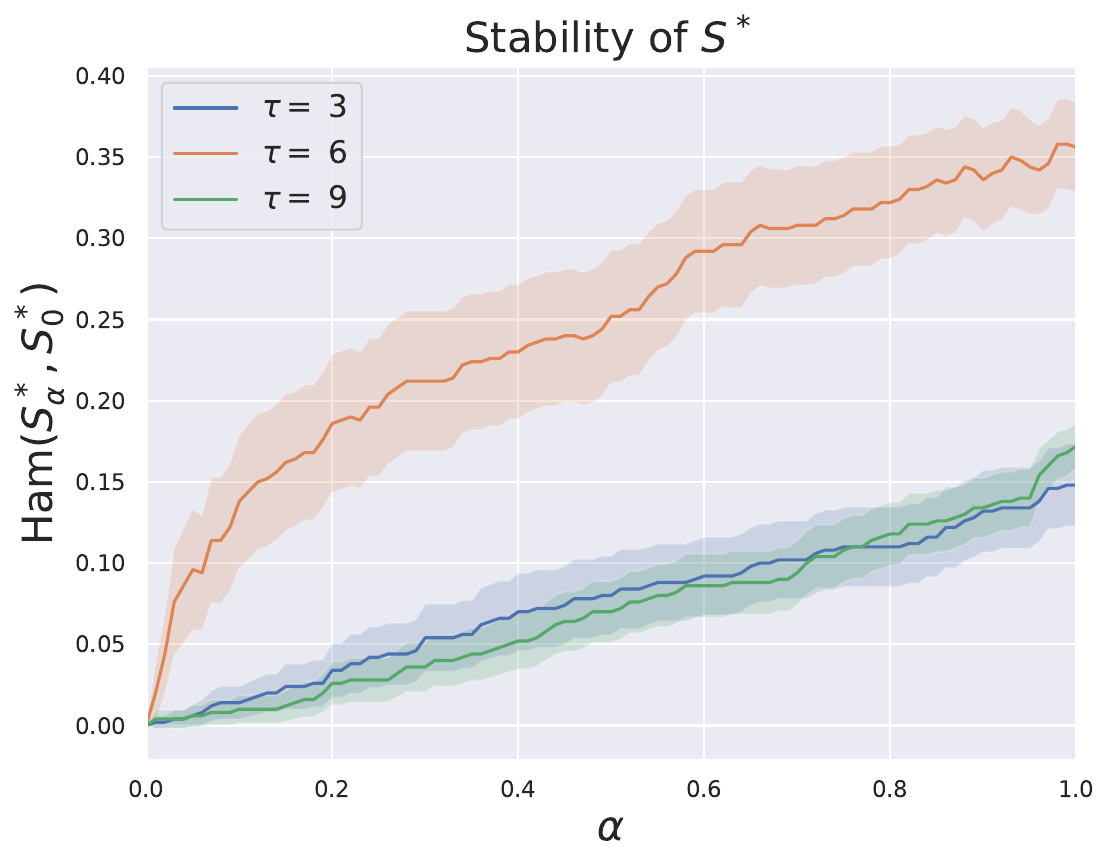}
        \caption{\hspace{0.2em}ACSIncome}
        \label{fig:acs}
    \end{subfigure}
    \label{fig:tabular}
    \caption{Stability of solutions to \eqref{eq:uni_opt} vs. $\alpha$ for $\tau \in \{3, 6, 9\}$}
\end{figure*}

\paragraph{Stability of Unified Solutions.} \cref{fig:acs} shows that when solutions are forced to be small (\(\tau = 3\)), increasing \(\alpha\) to enforce sufficiency results in a steady increase in Hamming distance, indicating the solutions \( S^*_{\alpha_i} \) are changing. For larger solutions (\(\tau = 6\)), \( S^*_{\alpha_i} \) changes significantly when low levels sufficiency are required, indicated by initial rise in the Hamming distance. 
As $\alpha$ continues to increase, the Hamming distance grows more gradually. Interestingly, when the size is close to feature space's dimensionality (\(\tau = 9\)), the Hamming distance exhibits a behavior similar to that observed for \(\tau = 3\). In conclusion, both experiments show that the optimal feature set can vary significantly depending on the size allowed and balance between sufficiency and necessity.

\subsection{Image Classification}
\vspace{-0.2cm}
The following two experiments explore high dimensional settings in image classification tasks. The features are pixel values and so a subset $S$ corresponds to a binary mask identifying important pixels. Since solving \eqref{eq:suff_opt}, \eqref{eq:necc_opt}, or \eqref{eq:uni_opt} is NP-hard, we use two methods--one for each setting, described in their respective sections--which solve relaxed problems to identify sufficient and necessary masks $S$. These experiments serve two purposes. First, they will analyze the extent to which explanations generated by popular methods--including Integrated Gradients \citep{sundararajan2017axiomatic}, GradientSHAP \citep{lundberg2017unified}, Guided GradCAM \citep{selvaraju2017grad}, and  h-Shap \citep{teneggi2022fast}--identify small sufficient and necessary subsets. To ensure consistent analysis, we normalize all generated attribution scores to the interval $[0,1]$. This is done by setting the top 1\% of nonzero scores to 1 and dividing the remaining scores by the minimum score from the top 1\% of nonzero scores. Then, binary masks are generated by thresholding the normalized scores using $t \in [0,1]$. For a test set of images, we perform this normalization and report the average $-\log(\Delta^{\suf})$, $-\log(\Delta^{\nec})$, and $-\log(L^0)$ (across all binary masks) at different threshold values to analyze the sufficiency, necessity and size of the explanations. Finally, the second  objective is to understand and visualize the similarities and differences between sufficient and necessary sets.

\subsubsection{RSNA CT Hemorrhage}
\vspace{-0.2cm}
We use the RSNA 2019 Brain CT Hemorrhage Challenge dataset comprised of 752,803 scans. Each scan is annotated by expert neuroradiologists with the presence and type(s) of hemorrhage (i.e., epidural, intraparenchymal, intraventricular, subarachnoid, or subdural). We use a ResNet18 \citep{he2016deep} classifier that was pretrained on this data \citep{teneggi2022fast}. To identify sufficient and necessary sets we solve the relaxed problem,
\begin{align}
    \label{eq:sample_exp_opt}
    \underset{S \subseteq [0,1]^d}{\text{arg min}} ~~ \Delta^{\uni}_\calV(S, f, \x, \alpha) + \lambda_{1}\cdot||S||_1 + \lambda_{\text{TV}}\cdot||S||_{TV}.
\end{align}
Since the dataset consists of highly complex and diverse images, we employ this per-example approach that generates highly specific tailored solutions by solving an optimization problem for each sample following previous work \citep{fong2019understanding, kolek2021cartoon, kolek2022rate}.\footnote{$\lambda_1, ~||S||_1$ and $\lambda_{\text{TV}}, ~ ||S||_{TV}$ are the $\ell_1$ and Total Variation norms and hyperparamters, promoting sparsity and smoothness.} To learn sufficient and/or necessary masks, we solve \cref{eq:sample_exp_opt} for \( \alpha \in \{0, 0.5, 1\} \). Details are in \cref{supp:exp_details}.

\begin{figure*}
    \begin{subfigure}{0.37\textwidth}
    \centering
    \includegraphics[width=5.3cm,height=9.1cm]{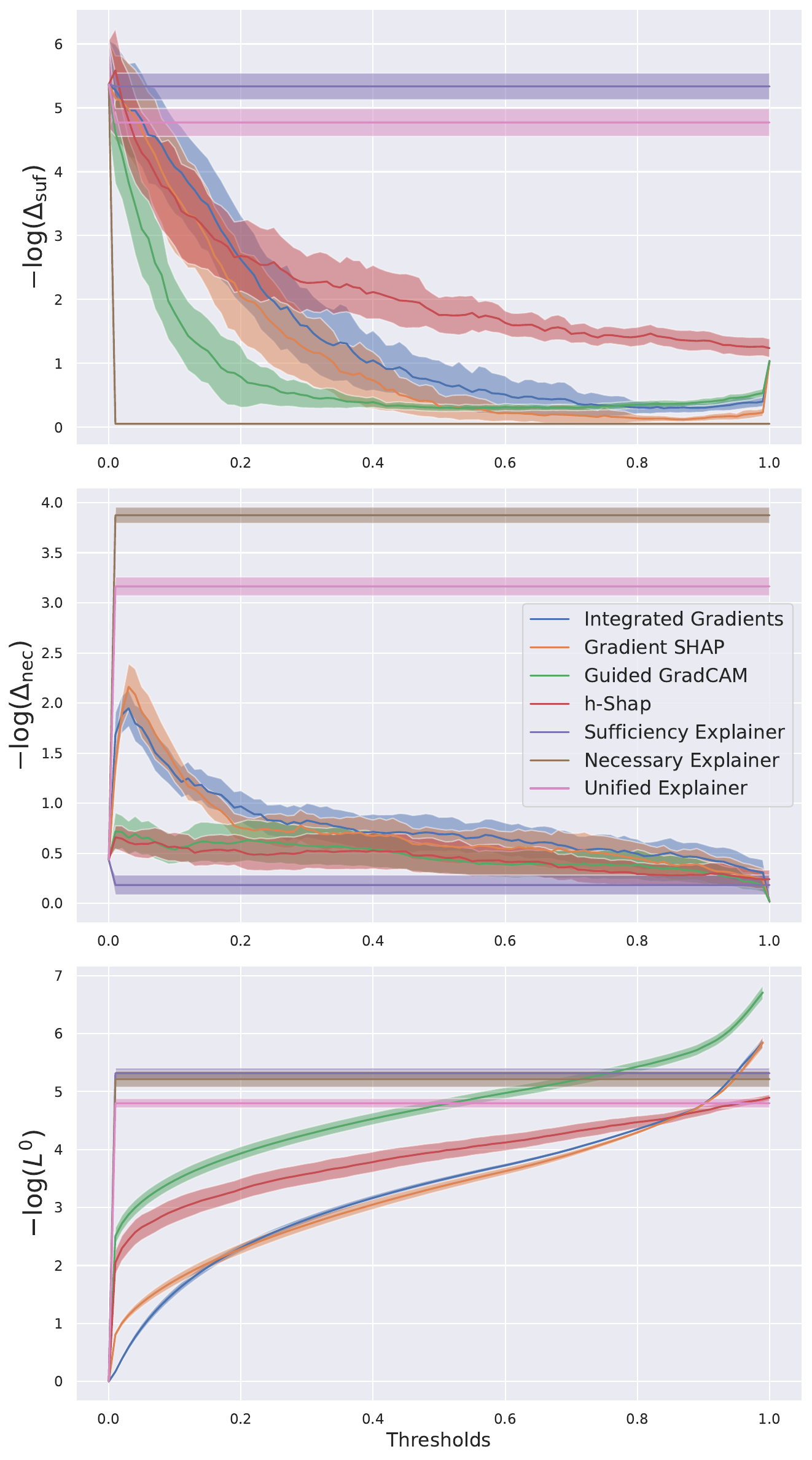}
        \caption{\hspace{0.2em} Comparison of different methods.}
        \label{fig:rsna_compare}
    \end{subfigure}
    \hfill
    \begin{subfigure}{0.62\textwidth}
        \centering
        \includegraphics[width=10.25cm,height=9.4cm]{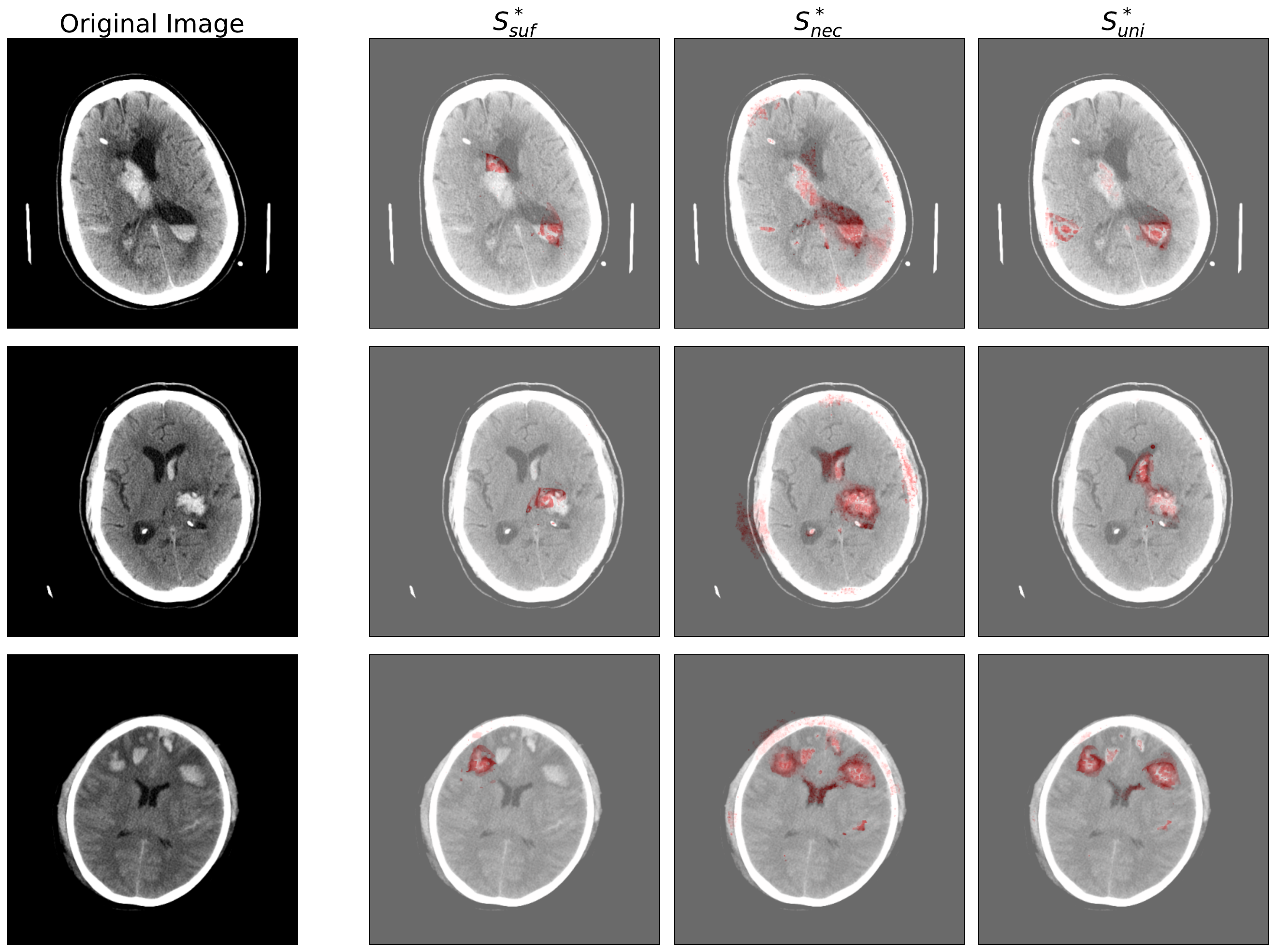}
        \caption{\hspace{0.2em}$S^*_{\suf}, ~S^*_{\nec}$ and $S^*_{\uni}$ for various CT scans.}
        \label{fig:rsna_suf_vs_nec}
    \end{subfigure}
    \label{fig:combined}
    \caption{Experimental results on the RSNA dataset.}
\end{figure*}

{\bf Comparison of Post-hoc Interpretability Methods.} For a set of 20 images positively classified by the ResNet model, we apply multiple post-hoc interpretability methods, as well as computing sufficient and necessary masks by our proposed approach -- solving \eqref{eq:sample_exp_opt}. The results in \cref{fig:rsna_compare} show that for a threshold of $t < 0.1$, many methods identify sufficient sets smaller in size than the sufficient and unified explainer, as indicated by their large values of $-\log(\Delta^{\suf})$ and smaller values of $-\log(L^0)$. However, for $t > 0.1$, only the sufficient and unified explainer identify sufficient sets of a constant small size. Importantly, it is also evident that \emph{no methods, besides our necessity and unified explainers, identify necessary sets}. Furthermore, as expected, the sufficient explainer does not identify necessary sets and vice versa. The unified explainer, as expected, identifies a sufficient and necessary set, albeit at the cost of the set being larger in size. In conclusion, while many methods can identify sufficient, no off-the-shelf method can identify necessary sets for small thresholds. Only when we directly optimize for such properties do we get explanations that are consistently small,  sufficient and/or necessary across thresholds.

{\bf Sufficiency vs. Necessity.}  
In \cref{fig:rsna_suf_vs_nec} we visualize the sufficient and necessary features in various CT scans. The first observation is that sufficient subsets do not provide a complete picture of which features are important. Notice for all the CT scans, a sufficient set, $S^*_{\suf}$ highlights one or two, but never all, brain hemorrhages in the scans. For example, in the last row, $S^*_{\suf}$ only contains the left frontal lobe parenchymal hemorrhages, which happens to be one of the larger hemorrhages present. On the other hand, necessary sets, $S^*_{\nec}$, contain parts of, sometimes entirely, \emph{all} hemorrhages in the scans. In the last row, $S^*_{\nec}$ contains all multifocal parenchymal hemorrhages in both right and left frontal lobes, because when all these regions are masked, the model yields a prediction $\approx 0.64$-- the prediction of the model on the mean image. Finally, notice in the 2nd and 3rd columns that $S^*_{\nec}$ and $S^*_{\uni}$ are nearly identical, which precisely demonstrate \cref{lemma:lemma1,theorem:thm1} in practice. First, since $S^*_{\suf}$ is super sufficient, $S^*_{\suf}$ and $S^*_{\nec}$, share common features. Second, visually $S^*_{\suf} \subseteq S^*_{\nec}$ holds approximately and so $S^*_{\nec} = S^*_{\uni}$. Through this experiment we are able to highlight the differences between sufficient and necessary sets, show how each contain important and complementary information, and demonstrate our theory holding in real world settings.

\subsubsection{CelebA-HQ}
We use a modified version of the CelebA-HQ dataset \citep{karras2017progressive} that contains 30,000 celebrity faces resized to 256$\times$256 pixels. We train a ResNet18 to classify whether a celebrity is smiling, achieving a test accuracy $\approx 94\%$. To generate sufficient or necessary masks $S$ for samples $\x$, we learn models $g_{\theta}: \calX \mapsto \calX$, that (approximately) solve the following optimization problem:
\begin{align}
    \label{eq:explainer_opt}
    \underset{\theta \in \Theta}{\text{arg min}}~\underset{\X \sim \mathcal{D}_{\mathcal{X}}}{\mathbb{E}}\left[\Delta^{\uni}_\calV(g_{\theta}(\X), f, \X, \alpha) + \lambda_{1}\cdot||g_{\theta}(\X)||_1 + \lambda_{\text{TV}}\cdot||g_{\theta}(\X)||_{\text{TV}}\right].
\end{align}
Given the structured nature of the dataset and the similarity of features across images, we use this parametric model approach because it prevents overfitting to spurious signals, an issue that can arise with per-example methods. Additionally, this approach is more efficient, as it still generates tailored per-sample explanations but only requires learning a single model rather than repeatedly solving \cref{eq:sample_exp_opt} \citep{linder2022interpreting, chen2018learning, yoon2018invase}. To learn a necessary and sufficient explainer model, we solve \cref{eq:explainer_opt} via empirical risk minimization for \( \alpha \in \{0, 1\}\) respectively. Implementation details and hyperparameter settings are included in \cref{supp:exp_details}.
\begin{figure*}
    \centering
    \includegraphics[width=16.5cm, height=3.75cm]{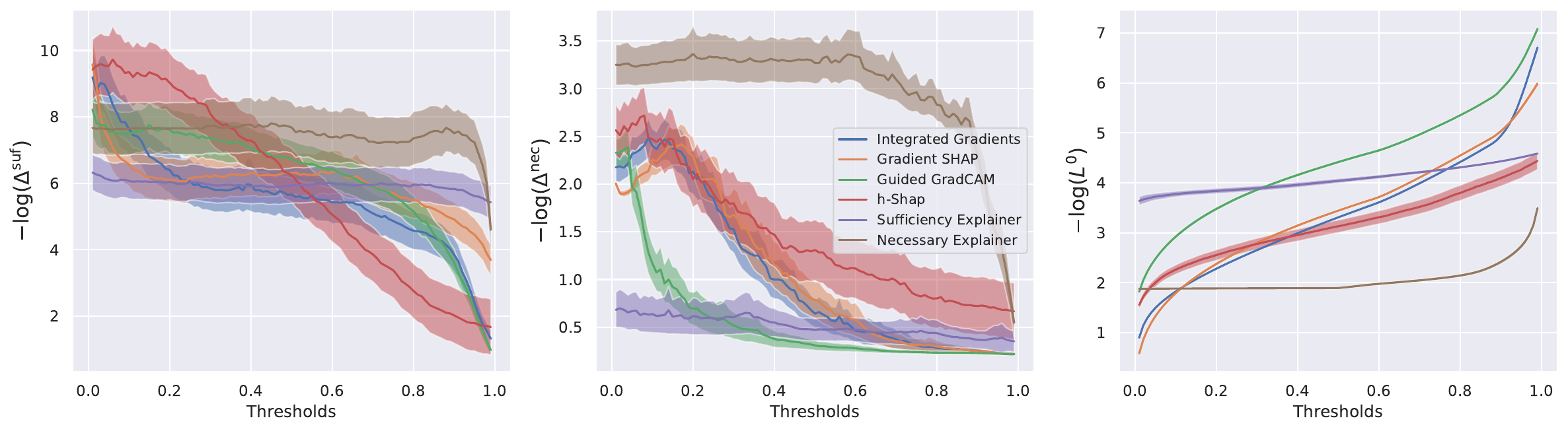}
    \caption{Comparison of different methods on the CelebAHQ dataset.}
    \label{fig:celebAcompare}
\end{figure*}

{\bf Comparison of Post-hoc Interpretability Methods.} For a set of 100 images labeled with a smile and correctly classified by the ResNet model, we apply multiple post-hoc interpretability methods and our sufficient and necessary explainers to identify important features associated with smiling. The results in \cref{fig:celebAcompare} illustrate that for a wide range of thresholds \( t \in [0,1] \), many methods identify sufficient subsets, as \( -\log(\Delta^{\suf}) \) for many of them is comparable to that of the sufficient explainer. The necessary explainer, in fact, identifies subsets that are more sufficient than those found by the sufficient explainer. The reason is that the sufficient explainer identifies subsets that are, on average, smaller for all \( t \in [0,1] \), while the necessary explainer finds subsets that are constant in size for all \( t \in [0,1] \) but slightly larger since, to be necessary, they must contain more features that provide additional information about the label. For other methods, as \( t \) increases, subset size decreases, and the sufficiency and necessity of the solutions decline. Meanwhile, the necessary explainer naturally identifies necessary subsets, indicated by large \( -\log(\Delta^{\nec}) \), whereas other methods fail to do so. In conclusion, many methods can identify sufficient sets, but not necessary ones. Directly optimizing for these criterion leads to identifying small, constant-sized subsets across thresholds.

\begin{figure*}
    \centering
    \includegraphics[width=0.9\textwidth]{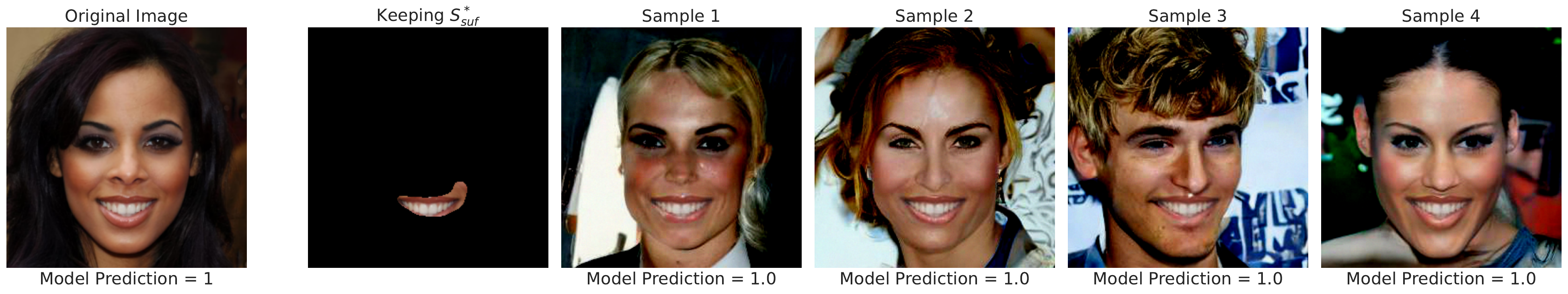}
    \includegraphics[width=0.9\textwidth]{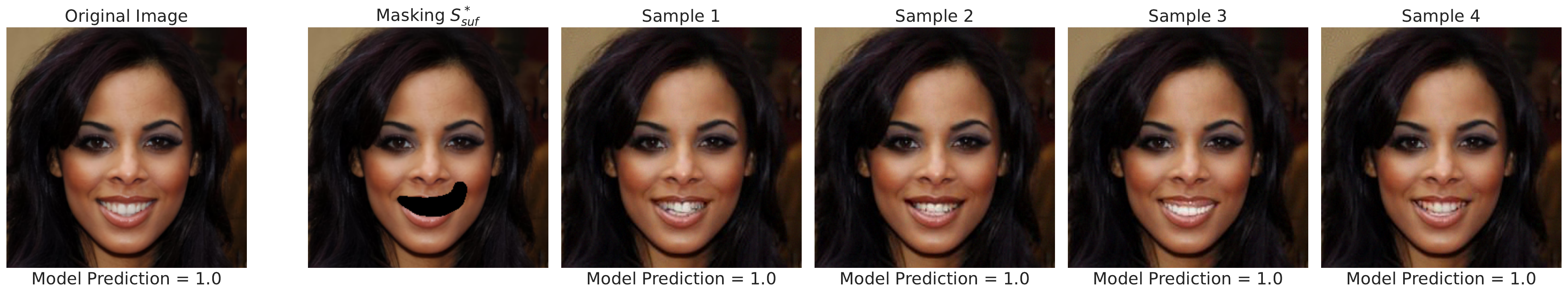}
    \caption{Images and model predictions by fixing and masking the sufficient subset $S^*_{\suf}$ }
    \label{fig:celebA_suff}
\end{figure*}

\begin{figure*}
    \centering
    \includegraphics[width=0.9\textwidth]{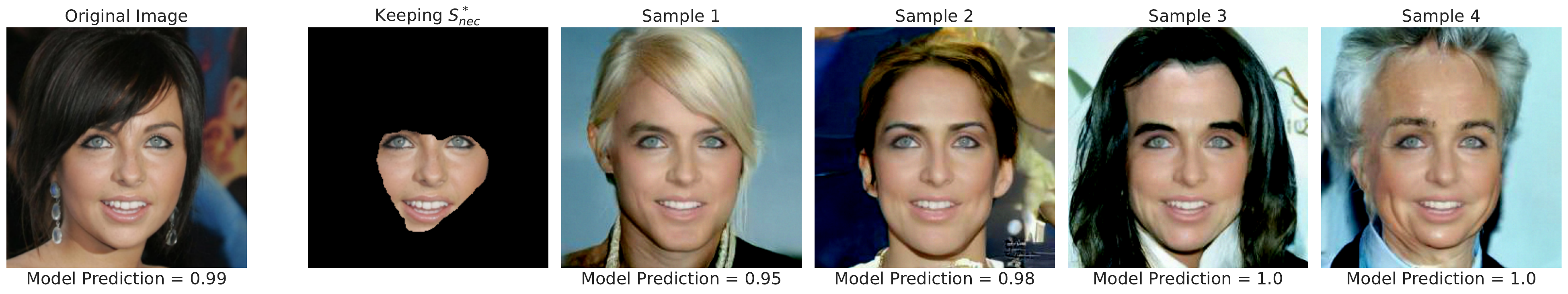}
    \includegraphics[width=0.9\textwidth]{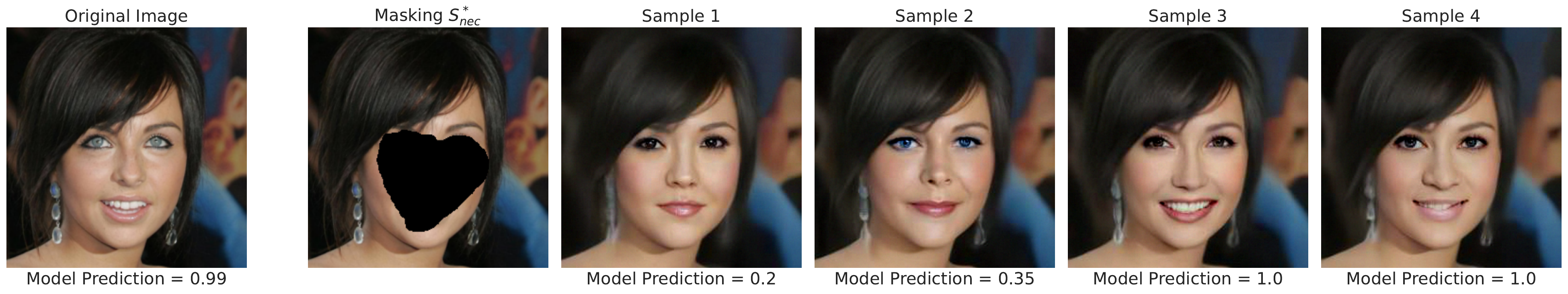}
    \caption{Images and model predictions by fixing and masking the necessary subset $S^*_{\nec}$}
    \label{fig:celebA_necc}
\end{figure*}

{\bf Sufficiency vs. Necessity.} In \cref{fig:celebA_suff}, we see how sufficient subsets alone may overlook important features, while solutions to \eqref{eq:uni_opt} offer deeper insights. As stated earlier, the sufficient explainer identifies sets that are sufficient but not necessary. On the other hand, the necessary explainer has high $-\log(\Delta^{\suf})$ and $-\log(\Delta^{\nec})$, indicating that it identifies sufficient \emph{and} necessary set, meaning they also serve as solutions to \eqref{eq:uni_opt}. In \cref{fig:celebA_suff}, we visualize the reasons for this phenomena. Notice that $S^*_{\suf}$ precisely highlights (only) the smile. When $S^*_{\suf}$ is fixed, one can generate new images (as done in \citep{zhang2023towards}) for which the model produces the same predictions as it did for the original image (a smile). On the other hand, we also see why $S^*_{\suf}$ is \textit{not} necessary: we can fix the complement $(S^*_{\suf})_c$ and, since there are important features in it, a smile is consistently generated, and the model produces the same prediction on these images as it did on the original. Conversely solutions to \eqref{eq:necc_opt} (also solutions to \eqref{eq:uni_opt} here) generate different explanations that provide a more complete picture of feature importance. Notice that $S^*_{\nec}$ is sufficient because $S^*_{\suf} \subseteq S^*_{\nec}$, with the additional features mainly being the dimples and eyes, which aid in determining the presence of a smile. More importantly, \cref{fig:celebA_necc} illustrates why $S^*_{\nec}$ is necessary: when we fix the complement of ${S}^*_{\nec}$ and generate new samples, half of the faces lack a smile, leading the model $f$ to predict no smile. Details on sample generation are in \cref{supp:proofs}.

\section{Limitations \& Broader Impacts}
While this work provides a novel theoretical contribution to the XAI community, there are some limitations that require careful discussion. The choice of reference distribution \(\mathcal{V}_S\) determines the characteristics of sufficient and necessary explanations. For instance, only with the true conditional data distribution can one obtain the conditional independence results that our theory provides. Naturally, there are computational trade-offs that must be carefully studied; the ability to learn and sample from accurate conditional distributions to generate explanations with clear statistical meaning comes with a computational and statistical cost, particularly in high-dimensional settings. Thus, a key direction for future work is to explore the impact of different reference distributions and provide a principled framework for selecting a \(\mathcal{V}_S\) that balances practical utility and computational feasibility. 

Another relevant question is how well our proposed notions align with human intuition. While we aim to understand which features are sufficient and necessary \emph{for a given predicted model}, these explanations may not always correspond to how humans perceive importance (since model might use different features to solve a task). This can be an issue in settings where interpretability is essential for trust and accountability, such as in healthcare. On the one hand, our approach can provide useful insights to further evaluate models (e.g. by verifying if the sufficient and necessary features employed by models correlate with the correct ones as informed by human experts). On the other hand, bridging the gap between our mathematical definitions of sufficiency and necessity and other human notions of importance is an area for further investigation. User studies, along with collaboration with domain experts, will be critical in determining how our formal notions of sufficiency and necessity can be adapted or extended to better meet real-world interpretability needs. 

Finally, the societal impact of this work warrants discussion. While we offer a rigorous framework to understand model predictions, 
these are oblivious to notions of demographic bias \citep{hardt2016equality, feldman2015certifying, bharti2024estimating}. There is a risk that an ``incorrect" choice of generating a sufficient vs. necessary explanation could reinforce biases or obscure the causal reasons behind predictions. Future work will study when and how our framework can be incorporate these biases in the reported important features.

\section{Conclusion}
This work formalizes notions of sufficiency and necessity as tools to evaluate feature importance and explain model predictions. We demonstrate that sufficient and necessary explanations, while insightful, often provide incomplete while complementary answers to model behavior. To address this limitation, we propose a unified approach that offers a new and more nuanced understanding of model behavior. Our unified approach expands the scope of explanations and reveals trade-offs between sufficiency and necessity, giving rise to new interpretations of feature importance. Through our theoretical contributions, we present conditions under which sufficiency and necessity align or diverge, and provide two perspectives of our unified approach through the lens of conditional independence and Shapley values. Our experimental results support our theoretical findings,  providing examples of how adjusting  sufficiency-necessity trade-off via our unified approach can uncover alternative sets of important features that would be missed by focusing solely on sufficiency or necessity. Furthermore, we evaluate common post-hoc interpretability methods showing that many fail to reliably identify features that are necessary or sufficient. In summary, our work contributes to a more complete understanding of feature importance through sufficiency and necessity. We believe, and hope, our framework holds potential for advancing the rigorous interpretability of ML models. 

\printbibliography

\newpage
\appendix
\section{Appendix}
\subsection{Proofs \label{supp:proofs}}
\subsubsection{Proof of \texorpdfstring{\cref{lemma:lemma1}}{Lemma 4.1}}
 \label{proof:lemma1}
{\bf Lemma 4.1.} Let $\alpha \in (0,1)$. For $\tau >0$, denote $S^*$ to be a solution to \eqref{eq:uni_opt} for which $\Delta^{\uni}_\calV(S^*, f, \x, \alpha)= \epsilon$. Then, $S^*$ is $\frac{\epsilon}{\alpha}$-sufficient and $\frac{\epsilon}{1-\alpha}$-necessary. Formally,
    \begin{align}
        0 \leq \Delta^{\suf}_\calV(S^*, f, \x) \leq \frac{\epsilon}{\alpha} \quad \text{and} \quad 0 \leq \Delta^{\nec}_\calV(S^*, f, \x) \leq \frac{\epsilon}{1-\alpha}.
    \end{align}
\begin{proof}
Let $\tau >0$ and $\alpha \in (0,1)$ and denote $S^*$ to be a solution to \eqref{eq:uni_opt} such that
\begin{align}
    \Delta^{\uni}_\calV(S^*, f, \x, \alpha) = \epsilon.
\end{align}
Then, by definition of being a solution to \eqref{eq:uni_opt},
\begin{align}
    |S^*| \leq \tau.
\end{align}
Furthermore, recall that
\begin{align}
\Delta^{\uni}_\calV(S^*, f, \x, \alpha) = \alpha\cdot \Delta^{\suf}_\calV(S^*, f, \x) + (1-\alpha)\cdot\Delta^{\nec}_\calV(S^*, f, \x)
\end{align}
which implies
\begin{align}
    \alpha\cdot \Delta^{\suf}_\calV(S^*, f, \x) &= \epsilon - (1-\alpha)\cdot\Delta^{\nec}_\calV(S^*, f, \x)\\ 
    &\leq \epsilon &~ ((1-\alpha), ~\Delta^{\nec}_\calV(S^*, f, \x) \geq 0)\\
    &\implies \Delta^{\suf}_\calV(S^*, f, \x) \leq \frac{\epsilon}{\alpha}.
\end{align}
Similarly,
\begin{align}
    (1-\alpha)\cdot \Delta^{\nec}_\calV(S^*, f, \x) &= \epsilon - \alpha\cdot\Delta^{\suf}_\calV(S^*, f, \x)\\ 
    &\leq \epsilon &~ (\alpha, ~\Delta^{\suf}_\calV(S^*, f, \x) \geq 0)\\
    &\implies \Delta^{\nec}_\calV(S^*, f, \x) \leq \frac{\epsilon}{1-\alpha}.
\end{align}
\end{proof}

\subsubsection{Proof of \texorpdfstring{\cref{lemma:lemma2}}{Lemma 4.2}}
 \label{proof:lemma2}
{\bf Lemma 4.2.} For $0 \leq \epsilon < \frac{\rho(f(\x), f_{\emptyset}(\x))}{2}$, denote $S^*_{\suf}$ and $S^*_{\nec}$ to be $\epsilon$-sufficient and $\epsilon$-necessary sets. Then, if $S^*_{\suf}$ is $\epsilon$-super sufficient or $S^*_{\nec}$ is $\epsilon$-super necessary, 
\begin{align}
    S^*_{\suf} \cap S^*_{\nec} \neq \emptyset.
\end{align}
\begin{proof}
We will prove the result via contradiction.  First recall that,
\begin{align}
    f_S(\x) = \underset{\X_{\cS} \sim \calV_{\cS}}{\E}[f(\x_S, \X_{\cS})] 
\end{align}
and, for any metric $\rho: \R \times \R \mapsto \R$,
\begin{align}
    \Delta^{\suf}_\calV(S, f, \x)  &\triangleq \rho(f(\x), f_{S}(\x))\\
    \Delta^{\nec}_\calV(S, f, \x)  &\triangleq \rho(f_{\cS}(\x), f_{\emptyset}(\x)).
\end{align}
Since $\rho$ is a metric on $\R$, it satisfies the triangle inequality. Thus, for $a,b,c \in \R$
\begin{align}
    \rho(a,c) \leq \rho(a,b) + \rho(b,c).
\end{align}
Now, let $S^*_{\suf}$ be $\epsilon$-super sufficient and suppose
\begin{align}
    S^*_{\suf} \cap S^*_{\nec} = \emptyset.
\end{align}
This implies
\begin{align}
    S^*_{\suf} \subseteq (S^*_{\nec})_c.
\end{align}
Subsequently, since $S^*_{\suf}$ is $\epsilon$-super sufficient,
\begin{align}
    \Delta^{\suf}_\calV((S^*_{\nec})_c, f, \x) \leq \epsilon.
\end{align}
As a result, observe
\begin{align}
    \rho(f(\x), f_{\emptyset}(\x)) &\leq \rho(f(\x), f_{(S^*_{\nec})_c}(\x)) + \rho(f_{(S^*_{\nec})_c}(\x), f_{\emptyset}(\x)) &~ \text{triangle inequality} \\
    &=   \Delta^{\suf}_\calV((S^*_{\nec})_c, f, \x) + \Delta^{\nec}_\calV((S^*_{\nec})_c, f, \x)\\
    &\leq \epsilon + \Delta^{\nec}_\calV((S^*_{\nec})_c, f, \x) &~ \text{$S^*_{\suf}$ is $\epsilon$-super sufficient}\\
    &\leq  2\epsilon &~ \text{$S^*_{\nec}$ is $\epsilon$-\text{necessary}}\\
    &\implies \epsilon \geq \frac{\rho(f(\x), f_{\emptyset}(\x))}{2}
\end{align}
which is a contradiction because $0 \leq \epsilon < \frac{\rho(f(\x), f_{\emptyset}(\x))}{2}$. Thus $S^*_{\suf} \cap S^*_{\nec} \neq \emptyset$. The proof of this result assuming $S^*_{\nec}$ is $\epsilon$-super necessary follows the same argument.
\end{proof}

\subsubsection{Proof of \texorpdfstring{\cref{theorem:thm1}}{Theorem 4.1}}
\label{proof:thm1}
{\bf Theorem 4.1.} Let $\tau_1, \tau_2 > 0$ and $0 \leq \epsilon < \frac{1}{2}\cdot\rho(f(\x), f_{\emptyset}(\x))$. Denote $S^*_{\suf}$ and $S^*_{\nec}$ to be $\epsilon$-super sufficient and $\epsilon$-super necessary solutions to \eqref{eq:suff_opt} and \eqref{eq:necc_opt}, respectively, such that $|S^*_{\suf}~| = \tau_1$ and $|S^*_{\nec}| = \tau_2$.
Then, there exists a set $S^*$ to \eqref{eq:uni_opt} such that 
\begin{align}
    \Delta^{\uni}_\calV(S^*, f, \x, \alpha) \leq \epsilon \quad \text{and} \quad
    \max(\tau_1, \tau_2) \leq |S^*| < \tau_1 + \tau_2.
\end{align}
Furthermore, if $S^*_{\suf} \subseteq S^*_{\nec}$ or $S^*_{\nec} \subseteq S^*_{\suf}$. then $S^* = S^*_{\nec}$ or $S^* = S^*_{\suf}$, respectively.
\begin{proof}
    Consider the set $S^* = S^*_{\suf} \cup S^*_{\nec}$. This set has the following properties:
    \begin{enumerate}[label=(P\arabic*)]
        \item $S^*$ is $\epsilon$-sufficient because $S^*_{\suf}$ is $\epsilon$-super sufficient
        \item $S^*$ is $\epsilon$-necessary because $S^*_{\suf}$ is $\epsilon$-super necessary
        \item $|S^*| \geq \max(\tau_1, \tau_2)$ with $|S^*| = \tau_1$ when $S^*_{\nec} \subset S^*_{\suf}$ and with $|S^*| = \tau_2$ when $S^*_{\suf} \subset S^*_{\nec}$
        \item Via \cref{lemma:lemma1}, we know $S^*_{\suf} \cap S^*_{\nec} \neq \emptyset$ thus $|S^*| < \tau_1 + \tau_2$
    \end{enumerate}
Then by (P1) and (P2)
\begin{align}
    \Delta^{\uni}_\calV(S^*, f, \x, \alpha) &= \alpha\cdot \Delta^{\suf}_\calV(S^*, f, \x) + 
    (1-\alpha)\cdot\Delta^{\nec}_\calV(S^*, f, \x)\\
    &\leq \alpha\cdot \epsilon + 
    (1-\alpha)\cdot\epsilon = \epsilon
\end{align}
and by (P3) and (P4) we have $\max(\tau_1, \tau_2) \leq |S^*| < \tau_1 + \tau_2$,
\end{proof}

\subsubsection{Proof of \texorpdfstring{\cref{corollary:corollary1}}{Corollary 5.1}}
{\bf Corollary 5.1.} Suppose for any $S \subseteq [d]$, $\calV_S = p({\X}_{S} \mid {\X}_{\cS} = {\x}_{\cS})$. Let $\alpha \in (0,1)$, $\epsilon \geq 0$, and denote $\rho: \R \times \R \mapsto \R$ to be a metric on $\R$. Furthermore, for $f(\X) = \E[Y \mid \X]$ and $\tau >0$, let $S^*$ be a solution to \eqref{eq:uni_opt} such that $\Delta^{\uni}_\calV(S, f, \x, \alpha) = \epsilon$. Then, $S^*$ satisfies the following conditional independence relations,
\begin{align}
    &\rho\left(\E[Y \mid \x],~ \E[Y \mid \X_{S^*} = \x_{S^*}]\right) \leq \frac{\epsilon}{\alpha} \quad \text{and} \quad \rho\left(\E[Y \mid \X_{S^*_c} = \x_{S^*_c}],~ \E[Y]\right) \leq \frac{\epsilon}{1-\alpha}.
\end{align}
\begin{proof}
    All we need to show is that when $\calV_S = p({\X}_{S} \mid {\X}_{\cS} = {\x}_{\cS})$ and $f(\X) = \E[Y \mid \X]$, we have
    \begin{align}
        f_{S}(\x) = \E[Y \mid {\bf X}_S = {\bf x}_S].
    \end{align}
    Once this is proven, we can simply apply \cref{lemma:lemma1}. 
    
    To this end, we have by assumption that $f(\x) = \E[Y \mid \X = \x]$ and, for any $S \subseteq [d]$, $\calV_S = p(\X_{S} \mid \X_{\cS} = \x_{\cS})$. Then by definition
    \begin{align}
        f_S(\x) = \E_{\calV_{\cS}}[f(\x_S,\X_{\cS})] &= \int_{\calX} f(\x_S,\X_{\cS}) \cdot p(\X_{\cS} \mid \X_{S} = \x_{S})~d\X_{\cS}\\
        &= \int_{\calX} \E[Y \mid \X_S = \x_S, \X_{\cS}] \cdot p(\X_{\cS} \mid \X_{S} = \x_{S})~d\X_{\cS}\\
        &= \int_\calX \left(\int_{\calY} y \cdot p(y \mid \X_S = \x_S, \X_{\cS})~dy\right) \cdot p(\X_{\cS} \mid \X_{S} = \x_{S})~d\X_{\cS}\\
        &= \int_\calY y \left(\int_{\calX} p(y, \X_{\cS}\mid \X_S = \x_S)~d\X_{\cS}\right)~dy\\
        &= \int_\calY y \cdot p(y \mid \X_S = \x_S)~dy\\
        &= \E[Y \mid \X_S = \x_S].
    \end{align}
    By applying \cref{lemma:lemma1}, we have the desired result.
\end{proof}

\subsubsection{Proof of \texorpdfstring{\cref{theorem:thm2}}{Theorem 6.1}}
{\bf Theorem 5.1.} Consider an input $\x$ for which $f(\x) \neq f_{\emptyset}(\x)$.  Denote by $\Lambda_d = \{S, \cS \}$ the partition of $[d] = \{1, 2, \dots, d\}$, and define the characteristic function to be $v(S) = -\rho(f(\x), f_{S}(\x))$. Then,
\begin{align}
   \phi^{\shap}_S(\Lambda_d, v) \geq \rho(f(\x), f_{\emptyset}(\x)) - \Delta^{\uni}_\calV(S, f, \x, \alpha).
\end{align}
\begin{proof}
    Before we prove the result, recall the following properties of a metric $\rho$ in the reals:
    \begin{enumerate}[label=(P\arabic*)]
        \item $\forall a, b \in \mathbb R, ~ \rho(a, b) = 0 \iff a=b$
        \item for $a,b,c \in \R, ~~~ \rho(a,c) \leq \rho(a,b) + \rho(b,c)$.
    \end{enumerate}
    
    Now, for the partition $\Lambda_d = \{S, \cS \}$ of $[d] = \{1, 2, \dots, d\}$ and characteristic function $v(S) = -\rho(f(\x), f_{S}(\x))$, $\phi^{\shap}_S(\Lambda_d, v)$ is defined as 
    \begin{align}
       \phi^{\shap}_S(\Lambda_d, v) &= \frac{1}{2}\cdot\left[v(S \cup \cS) - v(\cS)\right] + \frac{1}{2}\cdot\left[v(S) - v(\emptyset)\right]\\
       &= \frac{1}{2}\cdot\left[\rho(f(\x), f_{\cS}(\x)) - \rho(f(\x), f(\x))\right] + \frac{1}{2}\cdot\left[\rho(f(\x), f_{\emptyset}(\x)) - \rho(f(\x), f_{S}(\x))\right]\\
       &= \frac{1}{2}\cdot\left[\rho(f(\x), f_{\cS}(\x)) \right] + \frac{1}{2}\cdot\left[\rho(f(\x), f_{\emptyset}(\x)) - \rho(f(\x), f_{S}(\x))\right] \qquad \text{by (P1)}
    \end{align}
    By (P2)
    \begin{align}
        \rho(f(\x), f_{\emptyset}(\x)) &\leq \rho(f(\x), f_{\cS}(\x)) + \rho(f_{\cS}(\x), f_{\emptyset}(\x))\\
        &\implies \rho(f(\x), f_{\cS}(\x)) \geq  \rho(f(\x), f_{\emptyset}(\x)) - \rho(f_{\cS}(\x), f_{\emptyset}(\x)).
    \end{align}
    Thus 
    \begin{align}
        \phi^{\shap}_S(\Lambda_d, v) &= \frac{1}{2}\cdot\left[\rho(f(\x), f_{\cS}(\x)) \right] + \frac{1}{2}\cdot\left[\rho(f(\x), f_{\emptyset}(\x)) - \rho(f(\x), f_{S}(\x))\right]\\
        &\geq \frac{1}{2}\cdot\left[\rho(f(\x), f_{\emptyset}(\x)) - \rho(f_{\cS}(\x), f_{\emptyset}(\x)) \right] + \frac{1}{2}\cdot\left[\rho(f(\x), f_{\emptyset}(\x)) - \rho(f(\x), f_{S}(\x))\right]\\
        &= \rho(f(\x), f_{\emptyset}(\x)) - \Delta^{\uni}_\calV(S, f, \x, \alpha).
    \end{align}
\end{proof}

\subsection{Additional Experimental Details \label{supp:exp_details}}
In this section, we include further experimental details. All experiments were performed on a private cluster with 8 NVIDIA RTX A5000 with 24 GB of memory. All scripts were run on PyTorch \texttt{2.0.1}, Python \texttt{3.11.5}, and CUDA \texttt{12.2}.

\subsubsection{RSNA CT Hemorrhage}
\paragraph{Dataset Details.} The RSNA 2019 Brain CT Hemorrhage Challenge dataset \citep{flanders2020construction}, contains 752803 images labeled by a panel of board-certified radiologists with the types of hemorrhage present (epidural, intraparenchymal, intraventricular, subarachnoid, subdural).

\paragraph{Implementation.}
Recall for this experiment, to identify sufficient and necessary masks $S$ for a sample $\x$, we considered the relaxed optimization problem \citep{fong2019understanding, kolek2021cartoon, kolek2022rate}
\begin{align}
    \underset{S \subseteq [0,1]^d}{\text{arg min}} ~~ \Delta^{\uni}_\calV(S, f, \x, \alpha) + \lambda_{1}\cdot||S||_1 + \lambda_{\text{TV}}\cdot||S||_{TV}.
\end{align}
where $||S||_1$ and $||S||_{TV}$ are the $L^1$ and Total Variation norm of $S$, which promote sparsity and smoothness respectively and \( \lambda_{\text{Sp}} \) and \( \lambda_{\text{Sm}} \) are the associated. To solve this problem, a mask $S \in [0,1]^{512 \times 512}$ is initialized with entries $S_i \sim \calN(0.5, \frac{1}{36})$. For 1000 iterations, the mask $S$ is iteratively updated to minimize
\begin{align}
    \alpha \cdot |f(\x) - f_S(\x)| + 
    (1-\alpha) \cdot |f(\x) - f_S(\x)| + \lambda_{1}\cdot||S||_1 + \lambda_{\text{TV}}\cdot||S||_{TV}
\end{align}
where for any $S$, 
\begin{align}
    f_S(\x) = \frac{1}{K}\sum^K_{i=1}f((\tilde{\X}_S)_i)  \quad \text{with} \quad (\tilde{\X}_S)_i = \x \circ \tilde{\mathbbm{1}}_S + (1-\tilde{\mathbbm{1}}_S) \circ b_i.
\end{align}
Here the entries $(\tilde{\mathbbm{1}}_S)_i \sim \text{Bernoulli}(S_i)$ and $b_i$ is the $i$th entry of a vector ${\bf b} = (b_1, \cdots, b_d) \sim \calV$. In our implementation the reference distribution $\calV$ is the unconditional mean image over the of training images and so $b_i$ is the simply the average value of the $i$th pixel over the training set. To allow for differentiation during optimization, we generate discrete samples $\tilde{\mathbbm{1}}_S$ using the Gumbel-Softmax distribution. This methodology simply implies the entries $(\tilde{\X}_S)_i$ is a Bernoulli distribution with outcomes $\{b_i, x_i\}$, i.e. $(\tilde{\X}_S)_i$ is distributed as
\begin{align}
    &\Pr[(\tilde{\X}_S)_i = x_i] = S_i \\
    &\Pr[(\tilde{\X}_S)_i = b_i] = 1 - S_i
\end{align}
For each $\alpha \in \{0, 0.5, 1\}$, during optimization we set $K =10$, $\lambda_{1}=2$ and $\lambda_{\text{TV}}=20$ and use the Adam optimizer with default $\beta$-parameters of $\beta_1 = 0.9$, $\beta_2 = 0.99$ and a fixed learning rate of 0.1.

\subsubsection{CelebA-HQ}
\paragraph{Dataset Details.} We use a modified version of the CelebA-HQ dataset \citep{CelebAMask-HQ, karras2017progressive} which contains 30,000 celebrity faces resized to 256$\times$256 pixels with several landmark locations and binary attributes (e.g., eyeglasses, bangs, smiling).

\paragraph{Implementation.} Recall for this experiment, to generate sufficient or necessary masks $S$ for samples $\x$, we learn a model $g_{\theta}: \calX \mapsto [0,1]^d$ via solving the following optimization problem:
\begin{align}
    \underset{\theta \in \Theta}{\text{arg min}}~\underset{\X \sim \mathcal{D}_{\mathcal{X}}}{\mathbb{E}}\left[\Delta^{\uni}_\calV(g_{\theta}(\X), f, \X, \alpha) + \lambda_{1}\cdot||g_{\theta}(\X)||_1 + \lambda_{\text{TV}}\cdot||g_{\theta}(\X)||_{\text{TV}}\right]
\end{align}
To learn sufficient and necessary explainer models, we solve \cref{eq:explainer_opt} via empirical risk minimization for $\alpha \in \{0,1\}$ respectively. Given $N$ samples $\{\X_i\}^N_{i=1} \overset{\text{i.i.d.}}{\sim}  \calD_{X}$, we solve 
\begin{align}
    \frac{1}{N}\sum^N_{i=1}\left[\Delta^{\uni}_\calV(g_{\theta}(\X_i), f, \X_i, \alpha) + \lambda_{1}\cdot||g_{\theta}(\X_i)||_1 + \lambda_{\text{TV}}\cdot||g_{\theta}(\X_i)||_{\text{TV}}\right].
\end{align}
Here
\begin{align}
    \Delta^{\uni}_\calV(g_{\theta}(\x_i), f, \x_i, \alpha)=  \alpha \cdot |f(\x_i) - f_S(\x_i)| + 
    (1-\alpha) \cdot |f(\x_i) - f_S(\x_i)|
\end{align}
where is $f_S(\x_i)$ is evaluated in the same manner as in the RSNA experiment. For $\alpha = 0$,
$\lambda_{1}=0.1$ and $\lambda_{\text{TV}}=100$. For $\alpha = 1$,
$\lambda_{1}=1$ and $\lambda_{\text{TV}}=10$. For both $\alpha$, during optimization we use a batch size of 32, set $K=10$ and use the Adam optimizer with default $\beta$-parameters of $\beta_1 = 0.9$, $\beta_2 = 0.99$ and a fixed learning rate of $1 \times 10^{-4}$

{\bf Sampling.} To generate the samples in \cref{fig:celebA_suff,fig:celebA_necc}, samples we use the \texttt{CoPaint} method \citep{zhang2023towards}. We utilize their code base and pretrained diffusion models with the exact the same parameters as reported in the paper to perform conditional generation. Everything used is available at \href{https://github.com/UCSB-NLP-Chang/CoPaint}{https://github.com/UCSB-NLP-Chang/CoPaint}.

\end{document}